\documentclass{optica-article}

\journal{opticajournal} 

\articletype{Research Article}

\usepackage{lineno}
\usepackage{upgreek}
\usepackage{graphicx}
\usepackage{caption}
\usepackage{subfigure}
\usepackage{enumitem}
\usepackage[marginal]{footmisc} 

\usepackage{booktabs}   
\usepackage{array}      
\usepackage{graphicx}   
\usepackage{multirow}   

\usepackage{color}

\usepackage{soul}

\begin{document}

\title{Large-FOV RGBD Imaging via Structured PSF Coding in a Bio-inspired Monocentric System}

\author{Zongxi Yu,\authormark{1, \#} Xiaolong Qian,\authormark{1, \#} Shaohua Gao,\authormark{1} Qi Jiang,\authormark{1} Yao Gao,\authormark{1} Kailun Yang,\authormark{2,3} and Kaiwei Wang\authormark{1,*}}

\address{\authormark{1}the State Key Laboratory of Extreme Photonics and Instrumentation, Zhejiang University, Hangzhou 310027, China\\
\authormark{2} the School of Artificial Intelligence and Robotics, Hunan University, Changsha 410012, China \\
\authormark{3} the National Engineering Research Center of Robot Visual Perception and Control Technology, Hunan University, Changsha 410082, China\\
\authormark{\#} These authors contributed equally to this work.
}

\email{\authormark{*}wangkaiwei@zju.edu.cn}
\begin{abstract}
High-fidelity large-field-of-view (LFOV) 3D sensing, essential for autonomous platforms, is hindered by the coupling of anisotropic off-axis aberrations and the ill-posed nature of monocular depth estimation.
To address this fundamental physical bottleneck, we propose the Bio-inspired Monocentric Imaging (BMI) framework, a holistic co-design integrating a monocentric optical topology with a physics-aware reconstruction network.
By pairing a concentric spherical lens with a hemispherical sensor, our system structurally eliminates coma, astigmatism, and field curvature.
Instead of relying on external modulators or hardware redundancy, we exploit the intrinsic aberration as a deterministic, radially symmetric carrier for depth encoding.
Theoretical analysis via the Cramér-Rao Lower Bound confirms that this topology maintains superior depth sensitivity across the entire FOV.
To decode these optical cues, our framework utilizes a dual-head network to jointly recover high-fidelity All-in-Focus images and dense metric depth maps from single-shot captures.  
The efficacy of this co-design is evaluated through physically-based simulations of field-dependent PSFs across the full optical FOV.
On NYU Depth V2, our system achieves a superior balance of image fidelity ($31.15dB$ PSNR) and depth precision ($0.161m$ RMSE).
It further maintains consistent reconstruction under spatially varying PSFs sampled across a $120^\circ$ optical FOV, mitigating the peripheral degradation observed in conventional wide-field designs.
The source code is publicly available at \url{https://github.com/ZongxiYu-ZJU/BMI}.
\end{abstract}

\section{Introduction}
High-fidelity large-field-of-view (LFOV) 3D environmental awareness is a critical requirement for next-generation autonomous platforms, ranging from robotics to AR/VR headsets~\cite{zhang2016benefit,robinson2023robotic,fahlstrom2022introduction}.
However, achieving this capability within compact, lightweight form factors presents a fundamental physical trade-off. 
Traditional depth sensors like LiDAR~\cite{li2022progress} or stereo systems~\cite{chang2018pyramid} require bulky baselines or active emitters, hindering deployment in compact, large-FOV perception applications.
This renders monocular imaging the ideal compact alternative due to its structural simplicity and minimal spatial requirements, yet it is bound by strict optical limitations.
Specifically, conventional compact large-FOV lenses inevitably suffer from severe symmetry-breaking off-axis aberrations and field curvature. 
This leads to anisotropic and irregular degradation of the Point Spread Function (PSF) at the periphery, which fundamentally limits the fidelity of computational 3D reconstruction~\cite{welford2017aberrations}.
While data-driven Monocular Depth Estimation (MDE)~\cite{hu2024metric3d,bochkovskii2024depth,yang2024depth} attempts to infer depth from semantic priors, the problem remains physically ill-posed~\cite{hartley2003multiple}.
Crucially, in large-FOV imaging, the significant geometric distortion inherently decouples semantic features from physical depth, rendering pure algorithmic prediction unreliable~\cite{ai2022deep}. 
This necessitates a shift from purely digital inference to optical encoding, where depth information is physically embedded into the image structure before capture.

To address this ill-posedness, computational imaging employs PSF engineering to explicitly encode depth cues into the optical measurement. By modulating the wavefront, depth information is embedded into spatially varying blur kernels, rendering the recovery well-posed. 
Prominent strategies utilize specialized elements, including phase masks~\cite{wu2019phasecam3d,mel2024end}, Diffractive Optical Elements (DOEs)~\cite{zhuge2024calibration}, and more recently, split-aperture coding~\cite{shi2024split}, to synthesize complex, depth-sensitive PSFs.
However, extending these encoding strategies to large-FOV scenarios encounters a physical bottleneck characterized by disordered and anisotropic degradation. 
In conventional planar-sensor architectures, the optical response is strictly limited by the breakdown of isoplanatic assumptions~\cite{welford2017aberrations}.
At large field angles, symmetry-breaking off-axis aberrations, primarily astigmatism and coma, combine with severe pupil projection distortion (the $\cos\theta$ effect~\cite{sasian2019introduction}) to drastically disrupt the engineered PSF morphology.
This physical limitation renders precise phase modulation strategies inherently unstable at the periphery. 
For instance, the structured coding patterns in split-aperture~\cite{shi2024split} or DOE-based~\cite{zhuge2024calibration} designs rely on a consistent pupil geometry. 
Under extreme off-axis compression, these patterns undergo anisotropic deformation, leading to spatially variant and unpredictable encoding that decouples the blur cues from depth, making valid inference computationally intractable.
Consequently, relying on complex supplementary elements not only introduces manufacturing challenges, such as diffraction-induced energy loss and strict alignment tolerances~\cite{khonina2025advancements}, but more critically, fails to address the root cause of off-axis degradation. 
This underscores a pivotal research gap: there is a need for a robust large-FOV encoding methodology derived from the optical topology itself. 
Such a solution must harness rotationally symmetric aberrations to ensure spatially consistent physical encoding across the entire field, obviating the need for fragile external modulators.

To bridge this gap, we propose a geometry-driven solution inspired by the monocentric topology of aquatic vision~\cite{kim2020aquatic}. 
Instead of suppressing aberrations via complex aspherics, we exploit the system's strict rotational symmetry to impose a geometric constraint on the light field. 
By coupling this monocentric front-end with a hemispherical sensor, we structurally eliminate the field curvature and astigmatism. 
Consequently, irregular off-axis aberrations are replaced by a deterministic PSF evolution governed by radial field position and predictable pupil projection. The retained spherical aberration therefore serves as a continuous and field-consistent carrier of depth information across the evaluated FOV.

Building upon this structured optical front-end, we establish the Bio-inspired Monocentric Imaging (BMI) framework, a holistic co-design that synergizes a compact monocentric lens for radially symmetric depth encoding with a physics-aware dual-head network for joint depth and image recovery.
The system is validated through a rigorous full-field wave optics simulation that accounts for realistic sensor noise and diffraction. 
Furthermore, we provide a theoretical validation using Cramér-Rao Lower Bound (CRLB) analysis, confirming the superior depth sensitivity of our topology compared to conventional designs.
Comprehensive experiments on the NYU Depth V2 benchmark~\cite{silberman2012indoor} validate the superiority of this symmetry-induced encoding philosophy: our method achieves a depth RMSE of $0.161m$, significantly surpassing leading software-only MDE methods and outperforming recent deep optics counterparts in wide-field consistency. 
Concurrently, our system strikes a state-of-the-art balance between depth precision and image fidelity, evidenced by a PSNR score of $31.15dB$ and an SSIM score of $0.945$.

In summary, the contributions of this work are threefold. 
First, we propose a geometry-driven monocentric optical front-end that harnesses intrinsic aberration for depth encoding. 
Taking advantage of strict rotational symmetry, this design ensures the physical integrity of depth-dependent PSFs across a large FOV (demonstrated at $120^{\circ}$), transforming chaotic off-axis aberrations into consistent, radially predictable coding signals. 
Second, we construct a physically based forward model of spatially varying PSFs at field positions spanning the $120^\circ$ optical FOV, providing a controlled test bed for evaluating reconstruction under field-dependent optical degradation.
Third, we develop a physics-aware reconstruction network that effectively decouples the depth-induced blur from the field-dependent compressive deformation, enabling the precise joint recovery of scene geometry and texture. 
Collectively, this framework overcomes the peripheral performance collapse inherent to traditional planar architectures, achieving an excellent balance between All-in-Focus (AiF) image fidelity and metric depth precision at sampled field positions across the evaluated optical FOV.

\section{Related Work}
\noindent\textbf{Paradigms in Bio-inspired Camera Design.} 
Biological vision systems have evolved elegant solutions for large-FOV imaging. For instance, the human eye utilizes a curved retina to inherently compensate for field curvature~\cite{atchison2023optics,gao2022recent}, while aquatic species employ spherically symmetric lenses to achieve ultra-wide FOV up to $160^{\circ}$ ~\cite{he2024recent,jagger1999wide}.
Leveraging this aquatic morphology, we adopt a monocentric optical topology~\cite{kim2020aquatic}. Its global rotational symmetry inherently eliminates off-axis aberrations, ensuring consistent resolution and superior MTF across the entire field~\cite{zhang2022design}. 
To complement this, we incorporate a curved image sensor that physically compensates for Petzval field curvature~\cite{joaquina2023curved}. This design significantly reduces system volume while enhancing peripheral illumination and suppressing pixel crosstalk, thereby providing an ideal imaging plane for high-fidelity depth sensing~\cite{guenter2017highly}.

\noindent\textbf{PSF-Aware Depth Estimation.} 
The paradigm of depth acquisition has shifted from purely algorithmic monocular estimation to Deep Optics, where optical front-ends and reconstruction networks are jointly optimized. Classic approaches like Depth from Defocus (DfD) rely on native defocus blur~\cite{carvalho2018deep,wu2024self}, while modern frameworks employ specialized elements such as phase masks~\cite{wu2019phasecam3d,mel2024end} or freeform lenses~\cite{chang2019deep} to engineer depth-sensitive PSFs.

Similarly, DOEs have been adopted to encode depth information into the PSF structure. Baek~\textit{et al.}~\cite{baek2021single} designed a learned DOE to generate a PSF that varies with both depth and spectrum, enabling simultaneous single-shot hyperspectral and depth imaging. 
Ikoma~\textit{et al.}~\cite{ikoma2021depth} proposed a rotationally symmetric DOE and jointly trained the optics with a network using an occlusion-aware image formation model to achieve more accurate blur simulation at depth discontinuities. 
To address practical deployment challenges, Zhuge~\textit{et al.}~\cite{zhuge2024calibration} developed a calibration-free deep optics framework by combining ray tracing and diffraction to precisely simulate both on-axis and off-axis PSFs, thereby eliminating the need for physical system calibration.

Recently, Shi~\textit{et al.}~\cite{shi2024split} achieved single-shot RGBD via split-aperture coding, yet extending this to large FOV remains challenging as off-axis aberrations on planar sensors severely distort the coding patterns. Although Wei ~\textit{et al.}~\cite{wei2025learned} utilized off-aperture DOEs to mitigate these aberrations, diffractive elements inevitably introduce spectral dispersion and fabrication complexity. In contrast, our approach leverages the inherent rotational symmetry of a monocentric topology. 
By exploiting intrinsic spherical aberration rather than external masks, we ensure spatially efficient depth encoding across the optical FOV while avoiding the strongly wavelength-dependent phase modulation inherent to diffractive elements.

\noindent\textbf{Joint Depth Estimation and Image Restoration.} 
Unlike depth-aware deblurring~\cite{li2020dynamic,zhu2022deep,torres2024davide}, which relies on pre-existing depth maps, computational imaging systems encode depth into blur, necessitating a joint solution for depth and image recovery.
To address this, learning-based methods have evolved from interacting~\cite{gur2019single} and cascading strategies~\cite{anwar2021deblur} to shared-encoder dual-decoder architectures~\cite{nazir2023depth,hou2023joint} that exploit task correlations. However, these frameworks typically rely on generic camera models, overlooking the potential of optical front-end optimization to simplify the decoding process.

In contrast, we prioritize the geometric coherence of optical encoding. The monocentric topology produces a deterministic, radially structured PSF evolution, avoiding the irregular asymmetric aberrations typical of conventional wide-angle lenses. This regularity enables a shared-encoder dual-head MIMO-UNet~\cite{cho2021rethinking} to learn the field- and depth-dependent PSF manifold for joint image and depth recovery.

\section{Methodology}
\begin{figure}[!htb]
    \centering
    \includegraphics[width=1.0\linewidth]{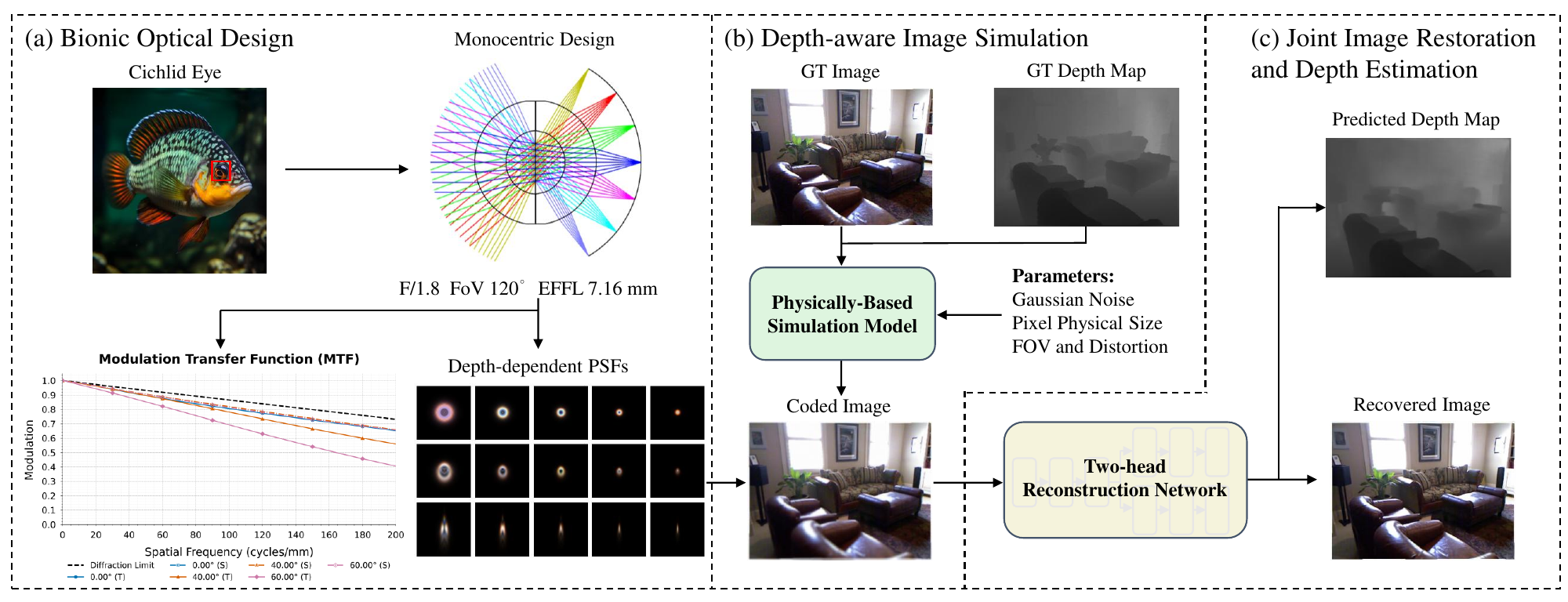}
    \caption{Overview of the Bio-inspired Monocentric Imaging (BMI) framework. (a) Optical Design: A monocentric topology generating rotationally symmetric, depth-dependent PSFs. (b) Simulation: An occlusion-aware forward model synthesizing coded captures via field-dependent pupil projection. (c) Reconstruction: A dual-head network decoding optical blur for joint AiF image and depth recovery.}
    \label{fig:overview}
\end{figure}

We propose the Bio-inspired Monocentric Imaging (BMI) framework, as illustrated in Fig.~\ref{fig:overview} which harnesses a strictly rotationally symmetric topology to deterministically encode depth via intrinsic aberration.
This section details the system architecture across three levels:
First, we define the optical design strategy (Sec.~\ref{section:optical design}) and provide a rigorous physical characterization of the aberration cues, validated by theoretical precision limits (Sec.~\ref{section:PSF analysis}).
Subsequently, we describe the physically-based forward simulation (Sec.~\ref{section:simulation}) that integrates field-dependent pupil projection and occlusion. 
Finally, we introduce the dual-head reconstruction network designed for single-shot joint recovery (Sec.~\ref{section:network}).

\subsection{Monocentric Optical Architecture and Aberration Analysis}
\label{section:optical design}
\noindent\textbf{Monocentric Geometry and Aberration Management.} 
The optical front-end is constructed based on a strict monocentric topology, inspired by the structural elegance of aquatic vision~\cite{kim2020aquatic}, employing a concentric arrangement of spherical glass elements sharing a common center of curvature.
This topological symmetry fundamentally redefines aberration management: since the optical axis is effectively defined by the chief ray at any incidence angle, asymmetric off-axis aberrations are structurally eliminated~\cite{welford2017aberrations}.
Specifically, coma is inherently absent due to the rotational symmetry relative to the chief ray, and astigmatism vanishes as the sagittal and tangential ray fans encounter identical surface curvatures.

The remaining primary aberrations are Petzval field curvature and spherical aberration.
To address the former, we incorporate a curved image sensor that geometrically conforms to the Petzval surface~\cite{rim2008optical}, ensuring focal plane consistency across the full $120^\circ$ FOV.
With field curvature and asymmetric aberrations eliminated, the optical response is dominated by Longitudinal Spherical Aberration (LSA).
Although LSA is theoretically field-invariant, the physical constraints of a finite aperture introduce a field-dependent geometric pupil projection.
We exploit this projected pupil foreshortening not as an artifact, but as a deterministic dual-parameter encoding strategy.
Since the pupil compression depends strictly on the field angle $\theta$, and the LSA depends primarily on object depth $d$, the resulting PSF evolves continuously within a predictable 2D manifold defined by $(d, \theta)$. This allows for robust depth decoding, as the spatial variation follows a deterministic radial pattern jointly resolvable with depth cues. The detailed design parameters are presented in Table~\ref{tab:fisheye}.

\begin{table}[!ht]\scriptsize
    \centering
       \caption{Bio-inspired optical design for the monocentric lens.}
       \label{tab:fisheye}  
\begin{tabular}{lcccc}  
\hline
Surface                    &  Radius (mm)&   Thickness (mm) &   Material & Semi-diameter (mm)\\ \hline
1 (Sphere)                 & 4.126     & 2.100  & H-ZLAF3    & 4.070   \\
2 (Sphere)                 & 2.103     & 2.000  & H-ZPK5     & 2.100   \\
3 (Stop)                   & infinity  & 2.000  & H-ZPK5     & 2.100   \\
4 (Sphere)                 & -2.103    & 2.100  & H-ZLAF3    & 2.100   \\
5 (Sphere)                 & -4.126    & 3.040  &-           & 4.070   \\
Sensor                     & -7.199    & -       &-          & 6.211 \\
\hline
\end{tabular}
\end{table}

\noindent\textbf{Analytical Modeling of Depth-Dependent Aberration.}
Following the structural elimination of asymmetric aberrations and field curvature (Sec.~\ref{section:optical design}), the residual wavefront error $W(\rho; d)$ is dominated by the superposition of rotationally symmetric defocus ($W_{020}$) and longitudinal spherical aberration ($W_{040}$)~\cite{sasian2019introduction}:
\begin{equation} 
{
W(\rho; d) = W_{020}(d) \cdot \rho^2 + W_{040}(d) \cdot \rho^4
}
\label{eq:wavefront_poly} 
\end{equation}
The defocus term $W_{020}$ serves as the primary dynamic depth indicator. 
Crucially, the system is designed with a fixed focus at optical infinity. 
This configuration enforces a strict monotonic evolution: as the object distance $d$ increases from the near field toward optical infinity, the geometric image plane converges toward the sensor surface, resulting in a continuous reduction in the magnitude of $W_{020}$.
This fundamentally eliminates the sign ambiguity inherent in traditional DfD methods. 
Simultaneously, the maximum defocus at the near limit remains moderate, ensuring that the image quality is not severely compromised even for close-range objects, thereby preserving sufficient structural information for effective restoration.

In contrast, the spherical aberration $W_{040}$ serves as a strong, quasi-static modulator. 
In our design, $W_{040}$ maintains a substantial positive magnitude ($\approx 2.5\lambda$ at the design wavelength $\lambda_d = 587.6$ nm) across the depth range. 
The interplay between the dynamic quadratic defocus ($W_{020}$) and this static quartic bias ($W_{040}$) transforms the PSF evolution from a simple scaling Gaussian into a structured topological transformation (\textit{e.g.}, from a ring pattern to a central peak). 
This characteristic structural variation significantly enhances the discriminability of depth cues and preserves high-frequency information, thereby enabling robust single-shot depth estimation.

\noindent\textbf{Field-Dependent PSF Formulation.}
While the monocentric topology preserves the rotational symmetry of aberration coefficients ($W_{020}, W_{040}$), the effective entrance pupil projection undergoes deterministic geometric compression at oblique angles. 
Due to the obliquity factor, the effective pupil area perpendicular to the chief ray is compressed in the tangential direction by $\cos\theta$, transforming the circular aperture into an ellipse. 
This projection breaks the radial symmetry of the PSF intensity distribution at the periphery.

We define the pupil plane coordinates as $(\xi, \eta)$, where $\eta$ aligns with the tangential field. The field-dependent binary pupil function $P(\xi, \eta; \theta)$ is modeled as an elliptical indicator function:
\begin{equation}
P(\xi, \eta; \theta) = \begin{cases} 1, & \text{if } \xi^2 + \left(\frac{\eta}{\cos\theta}\right)^2 \leq R^2 \\ 
0, & \text{otherwise} \end{cases},
\end{equation}
where $R$ is the physical aperture radius. By combining this amplitude modulation with the depth-dependent phase modulation derived in Eq.~\ref{eq:wavefront_poly}, the spatially variant PSF is derived via the Fourier transform of the generalized pupil function:
\begin{equation}
\text{PSF}(u, v; d, \theta) = \left| \mathcal{F} \left\{ P(\xi, \eta; \theta) \cdot \exp\left( j \frac{2\pi}{\lambda} \left[ W_{020}(d)\rho^2 + W_{040}(d)\rho^4 \right] \right) \right\} \right|^2,
\label{eq:final_psf_model}
\end{equation}
where the normalized radial coordinate corresponds to $\rho = \sqrt{\xi^2 + \eta^2}/R$. 
This formulation explicitly describes our dual-parameter encoding strategy: 
the phase term $W(\rho; d)$ governs the topological structure (\textit{e.g.}, evolution from rings to cores) based on depth $d$, while the pupil function $P$ governs the geometric envelope (eccentricity) based on field angle $\theta$. This functional orthogonality enables the reconstruction network to effectively disentangle depth-induced defocus cues from field-induced projection distortions.

\subsection{Physical Characterization and Theoretical Precision Limits}
\label{section:PSF analysis}

\noindent\textbf{Physical Characterization of Aberration Cues.}
To verify the aberration consistency modeled in Sec.~\ref{section:optical design}, we performed a full-field wave optics simulation of the proposed monocentric system.
Unlike conventional wide-FOV lenses, which suffer from irregular symmetry-breaking aberrations at the periphery, our topology guarantees that the optical response evolves in a strictly deterministic manner.
The primary off-axis variation arises solely from the geometric projection of the entrance pupil, which transitions from circular to elliptical following a $\cos\theta$ scaling law.

\begin{figure}[!htb]
    \centering
    \includegraphics[width=1.0\linewidth]{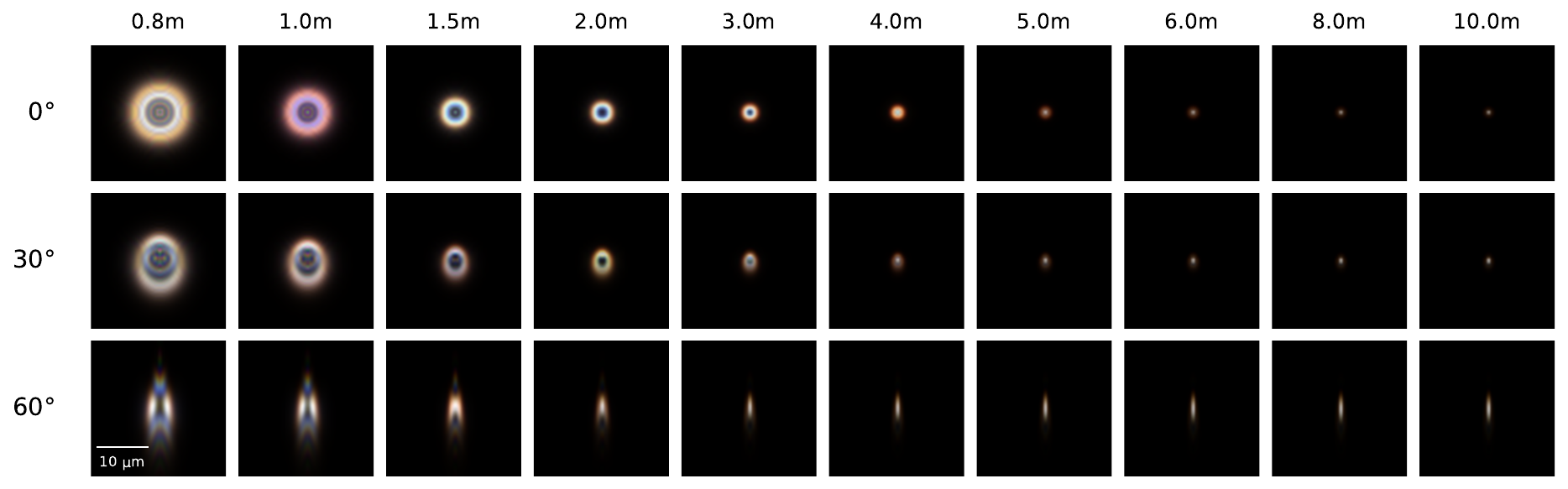}
    \caption{Simulated PSF evolution across the HFOV ($0^\circ$-$60^\circ$) and depth range ($0.8m$-$10m$). The results exhibit deterministic anisotropic compression at oblique angles while preserving the depth-dependent topological structure (invariant ring to core transition) across the entire field.}
    \label{fig:PSFs}
\end{figure}

As visualized in Figure~\ref{fig:PSFs}, this projection introduces a predictable anisotropic compression effect to the PSF morphology but preserves its inherent topological structure derived from spherical aberration. 
This behavior supports two critical conclusions for depth perception.
First, the aberration rings exhibit high depth sensitivity, functioning as a continuous scaling metric that evolves distinctly with object distance.
Second, even at the extreme $60^\circ$ field, the intensity distribution remains topologically isomorphic to the on-axis response.
This confirms that the depth code is spatially invariant in topology, merely subject to a known affine transformation, which allows the reconstruction network to apply a unified decoding logic across the entire $120^\circ$ FOV.

\noindent\textbf{Theoretical Limit Analysis via CRLB.}
To rigorously quantify the intrinsic depth encoding capability of our optical front-end, decoupled from downstream reconstruction algorithms, we analyze the theoretical lower bound of depth estimation variance. 
We invoke the Cramér-Rao Lower Bound (CRLB) theorem~\cite{kay1993fundamentals}, which establishes that the variance of any unbiased depth estimator $\hat{d}$ is bounded by the inverse of the Fisher Information $\mathcal{I}(d)$, expressed as:
\begin{equation}
\text{Var}(\hat{d}) \ge \mathcal{I}(d)^{-1}.
\end{equation}

Assuming standard illumination conditions where Poissonian-Gaussian sensor noise effectively converges to a Gaussian process~\cite{foi2008practical}, we model the sensor measurement as the convolution of the latent scene $s$ and the depth-dependent PSF, perturbed by additive Gaussian noise with variance $\sigma^2$.
Consequently, the Fisher Information $\mathcal{I}(d)$ is defined as~\cite{kay1993fundamentals}:
\begin{equation} 
\mathcal{I}(d) = \frac{1}{\sigma^2} \sum_{i} \left( \frac{\partial S_i(d)}{\partial d} \right)^2 
= \frac{1}{\sigma^2} \sum_{x,y} \left( \frac{\partial (s * \mathrm{PSF})}{\partial d} \right)^2 .
\end{equation}

Since the latent scene $s$ is depth-invariant, the derivative applies solely to the kernel: $\frac{\partial (s * PSF)}{\partial d} = s * \frac{\partial PSF}{\partial d}$. By invoking Parseval's theorem, we transform the spatial energy summation into the frequency domain to separate the scene spectrum from the optical transfer function:
\begin{equation} 
\sum_{x,y} \left( s \ast \frac{\partial PSF}{\partial d} \right)^2 = \sum_{u,v} \left| \mathcal{F} \left\{ s \ast \frac{\partial PSF}{\partial d} \right\} \right|^2 .
\end{equation}

By applying the convolution theorem,  the Fourier transform of the spatial convolution decomposes into the element-wise product of the scene spectrum $S(u,v)$ and the derivative of the Optical Transfer Function, denoted as $\frac{\partial \mathrm{OTF}(u,v; d)}{\partial d}$. Substituting this frequency-domain representation back into the Fisher Information formulation explicitly separates the texture component from the optical sensitivity component:

\begin{equation}
\mathcal{I}(d) = \frac{1}{\sigma^2} \sum_{u,v} \left| S(u,v) \cdot \frac{\partial OTF(u,v; d)}{\partial d} \right|^2 = \frac{1}{\sigma^2} \sum_{u,v} |S(u,v)|^2 \cdot \left| \frac{\partial OTF(u,v; d)}{\partial d} \right|^2.
\label{eq:fenli}
\end{equation}

\begin{figure}[!b]
    \centering
    \includegraphics[width=0.8\linewidth]{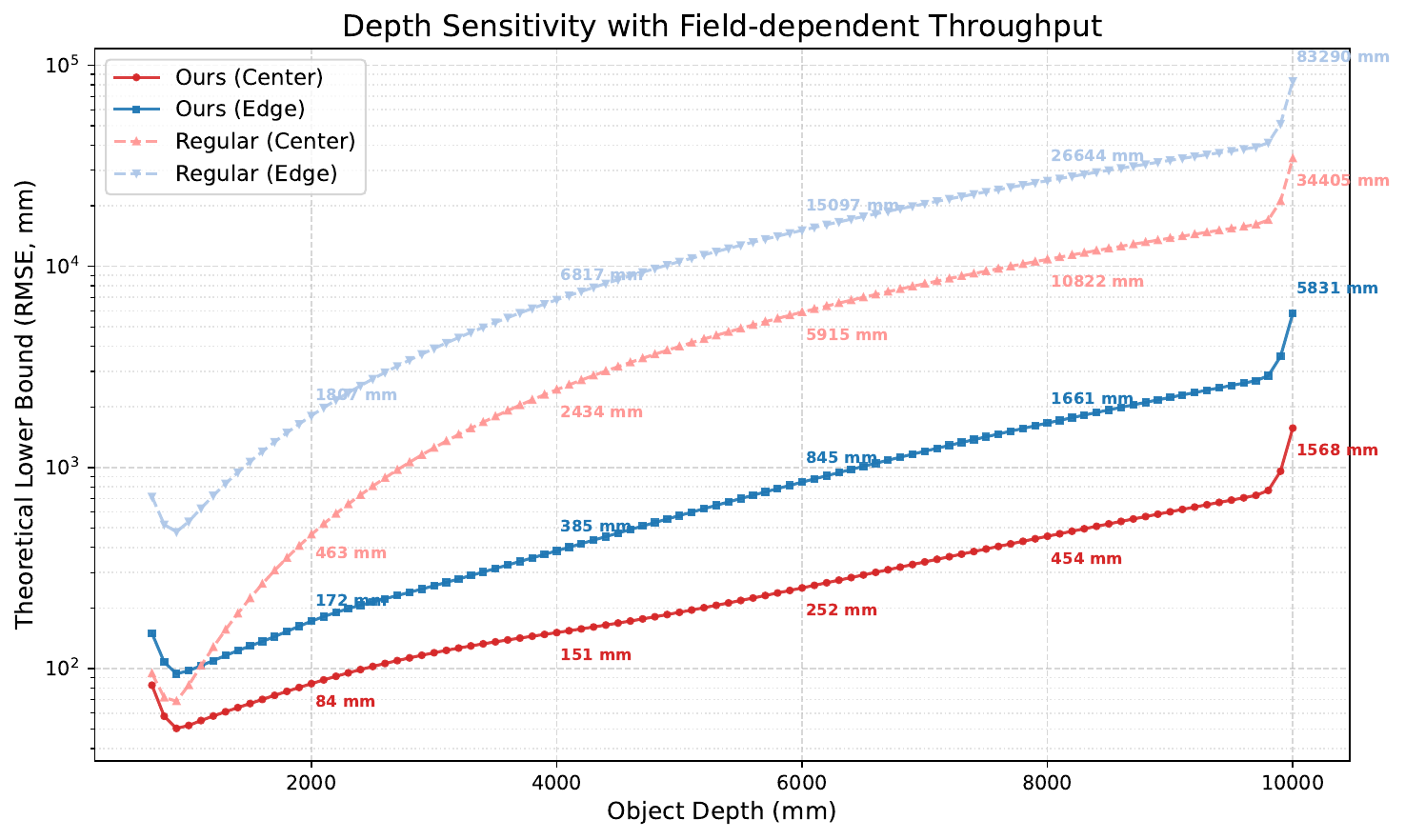}
    \caption{Theoretical precision limit (CRLB) comparison between the proposed bionic monocentric system and a conventional fisheye lens with field-dependent throughput.}
    \label{fig:CRLB}
\end{figure}

To numerically evaluate the CRLB, we use a natural image spectrum proportional to $1/f^{0.8}$ and a fixed Gaussian noise floor corresponding to quantization at 8 bits. The complex OTF at each depth is scaled by the throughput function $T(\theta)=\cos\theta$, which accounts for light loss caused by oblique pupil projection under a fixed exposure. 
Each color-channel PSF is independently normalized to unit energy, and the
Fisher information is evaluated from the squared magnitude of the complex-OTF
depth derivative in Eq.~\ref{eq:fenli}, thereby separating channel-wise PSF morphology
from the field-dependent throughput.

At $4.0~\mathrm{m}$, the proposed system yields theoretical depth bounds of $151~\mathrm{mm}$ at the center and $385~\mathrm{mm}$ at a field angle of $60^\circ$, compared with $2434~\mathrm{mm}$ and $6817~\mathrm{mm}$ for the fisheye baseline. At $6.0~\mathrm{m}$, the corresponding bounds are $252~\mathrm{mm}$ and $845~\mathrm{mm}$ for our system, versus $5915~\mathrm{mm}$ and $15097~\mathrm{mm}$ for the fisheye. Although the reduced throughput causes the expected degradation near the field edge, the monocentric design retains substantially stronger theoretical depth sensitivity across the evaluated field.

These CRLB values are local precision bounds for unbiased estimators under the adopted scene spectrum and Gaussian noise assumptions. They need not coincide with neural reconstruction errors because learned estimators may also exploit spatial context and scene priors.

\subsection{Physically-Based Forward Model and Dataset Simulation}
\label{section:simulation}
\noindent\textbf{PSF map construction.}
To bridge optical wavefront engineering with the computational reconstruction pipeline, we explicitly modeled the system's spatially varying response using ZEMAX~\cite{zemax_opticstudio_2023}.
Leveraging the strict rotational symmetry of the monocentric topology, we simulated a dense PSF tensor $\text{PSF}(c, \theta, d)$ spanning a depth range of 0.7 m to 10.0 m (0.1 m interval) across the full $120^{\circ}$ FOV ($0.6^{\circ}$ angular step).
Crucially, the simulation incorporates a curved image surface geometrically matched to the Petzval curvature, ensuring structural consistency of the PSF across the wide field and preventing the peripheral degradation typical of planar sensors.
Given this global axisymmetry, the full spatially variable PSF map is efficiently reconstructed from the radial tensor.
Following~\cite{gao2025exploring}, we employ interpolation, rotation, and resizing to synthesize the continuous PSF manifold for any arbitrary scene point $(h,w)$:

\begin{equation}
{\rm{PSF}}_{map}(c,h,w,d) = P_{\rm{resize}} \circ P_{\rm{rot}}(\sum\limits_\theta W(\theta) \cdot {\rm{PSF}}(c,\theta,d)),
\end{equation}
where $P_{resize}$ aligns the sampling pitch with the pixel size, $P_{rot}$ applies the field-dependent coordinate rotation, and $W(\theta)$ represents interpolation weights determined by an inverse square law.

\noindent\textbf{Depth-aware Image Simulation with Occlusion.} 
Conventional approaches typically rely on a locally isoplanatic approximation, partitioning the image into patches and convolving each with a uniform PSF~\cite{qian2025towards}. 
However, this simplification introduces discretization artifacts and fails to model the continuous blur gradients inherent to wide-field systems, particularly at depth discontinuities.
To ensure high-fidelity physical realism, we implement an occlusion-aware image formation model that integrates both the radial position $(h, w)$ and the object depth $d$ for every scene point~\cite{ikoma2021depth}. 
The scene is discretized into $K$ distinct depth layers. For each layer $k$, the image content $I_k$ and its corresponding alpha mask $\alpha_k$ are convolved with a depth-specific and spatially-varying PSF derived from our monocentric manifold. By incorporating the wavelength-dependent response $c$, the finalized image $P(c)$ is expressed as:

\begin{equation}
P(c) = \sum_{k=0}^{K-1} \tilde{I}_k \prod_{k'=k+1}^{K-1} (1 - \tilde{\alpha}_{k'}) + \eta,
\end{equation}
where $\tilde{I}_k = \frac{PSF_k(c) * I_k}{E_k(c)}$ and $\tilde{\alpha}_k = \frac{PSF_k(c) * \alpha_k}{E_k(c)}$. 
The term $\eta$ represents additive Gaussian noise, and $*$ denotes the convolution operator. 
The normalization term $E_k(c) {=} PSF(c) * \sum_{k'=0}^{k} \alpha_{k'}$ preserves energy conservation across layers.
This pixel-wise modeling of the PSF evolution enables the BMI framework to decode subtle structural cues embedded in the blur, which is crucial for achieving consistent reconstruction under spatially varying PSFs at field positions spanning the full $120^\circ$ optical FOV.
For the baseline spectral model, PSFs are simulated at representative wavelengths of $656.3$, $587.6$, and $486.1~\mathrm{nm}$ for the R, G, and B channels, respectively. These wavelengths capture the variation among the three color channels. Integration over the spectral response within each channel is evaluated separately using 13 wavelengths and published representative RGB response curves.
The depth range from $0.7$ to $10.0~\mathrm{m}$ is divided into $K=94$ layers at $0.1~\mathrm{m}$ intervals.

\subsection{Joint Image Restoration and Depth Estimation Network}
\label{section:network}
\begin{figure}[!ht]
    \centering
    \includegraphics[width=1.0\linewidth]{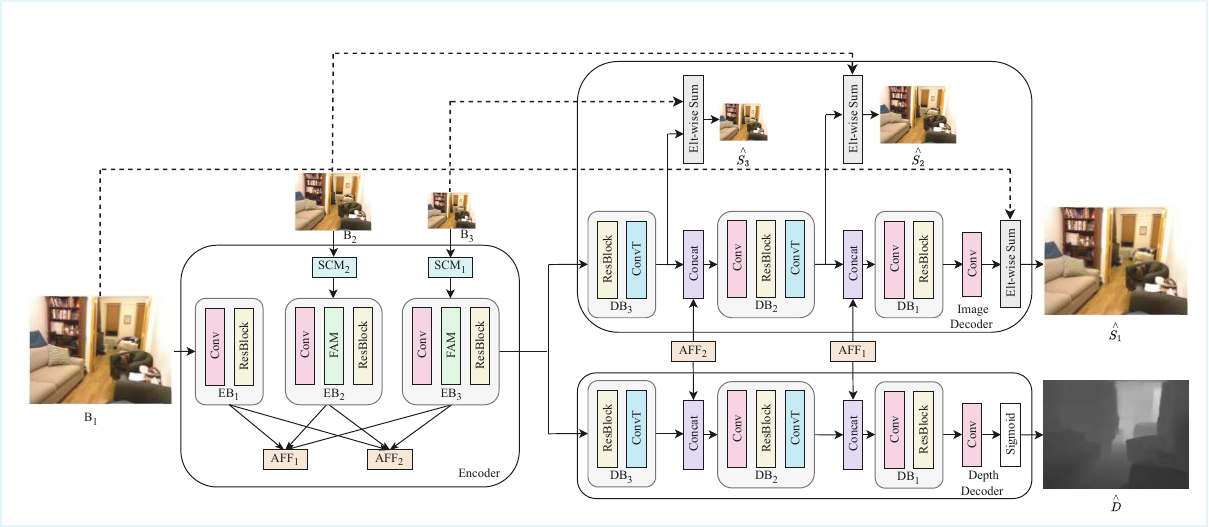}
    \caption{Architecture of the dual-head reconstruction network: a shared encoder extracts unified features from multi-scale coded inputs, and two parallel decoders simultaneously recover high-fidelity AiF images and metric depth maps.}
    \label{fig:network}
\end{figure}

Given the intrinsic coupling of depth information and image degradation within the monocentric PSF, we formulate the reconstruction task as simultaneous depth extraction and image restoration from a single frame.
We adapt a multi-scale image restoration network, MIMOUNet~\cite{cho2021rethinking} into a dual-head architecture~\cite{hou2023joint}, explicitly designed to discriminate depth by recognizing the distinct PSF patterns associated with different depth layers.
As illustrated in Fig.~\ref{fig:network}, our network adopts a multi-scale input strategy. 
The coded image $B_1$ is downsampled to generate $B_2$ ($\times 2$) and $B_3$ ($\times 4$), with all scales feeding into a shared encoder.
The coarse-scale inputs enlarge the effective receptive field for recognizing PSF structures, whereas the full-resolution input preserves fine textures and object boundaries. This enables the shared encoder to jointly capture global depth-encoding morphology and local image details.
The architecture features two branches for image restoration and depth estimation, which share a common encoder and feature fusion modules. 
This design ultimately produces multi-scale, all-in-focus image outputs and a final depth map prediction. This joint learning paradigm effectively internalizes the radial symmetry prior of the monocentric lens, allowing a unified network to handle field-dependent PSF variations across sensor positions spanning the entire optical FOV.
The training loss function $L_{total}$ consists of image loss $L_{cont}$ and $L_{MSFR}$ as well as depth loss $L_{Silog }$ as: 
\begin{equation}
{L_{total}} = {\gamma _{cont}}{L_{cont}} + {\gamma _{MSFR}}{L_{MSFR}} + {\gamma _{silog }}{L_{silog }}.
\end{equation}

For image restoration, we employ a multi-scale content loss $L_{cont}$ (Eq.~\ref{eq:content_loss}), calculated as the $L_1$ distance between restored and ground-truth images at each scale ~\cite{nah2017deep}.
To explicitly enforce high-frequency recovery, we incorporate a multi-scale frequency-domain loss $L_{MSFR}$ (Eq.~\ref{eq:freq_loss}), which computes the $L_1$ distance between their Fourier transforms~\cite{jiang2021focal}.
\begin{equation}
{L_{cont}} = \sum\limits_{k = 1}^K {\frac{1}{{{t_k}}}} {\left\| {{{\hat S}_k} - {S_k}} \right\|_1}.
\label{eq:content_loss}
\end{equation}
\begin{equation}
L_{MSFR} = \sum_{k=1}^{K} \frac{1}{t_k} \left\| \mathcal{F}(\hat{S}_k) - \mathcal{F}(S_k) \right\|_1.
\label{eq:freq_loss}
\end{equation}

For depth estimation, we utilize the scale-invariant log error loss $L_{Silog}$~\cite{eigen2014depth}. By penalizing relative depth inconsistencies while ignoring absolute global scale, this metric significantly enhances training stability. $L_{Silog}$ is defined as:
\begin{equation}
{L_{Silog }} = \frac{1}{N}\sum\limits_{i = 1}^N {(\log {d_i} - \log {{\hat d}_i})^2 - \frac{1}{{{N^2}}}}(\sum\limits_{i = 1}^N {{{(\log {d_i} - \log {{\hat d}_i})})^2}}.
\end{equation}

\section{Experiments and Analyses}
\subsection{Datasets and Implementation Details}
\noindent\textbf{Dataset selection.} 
To ensure robust generalization and prevent overfitting, we implement a hybrid strategy combining the NYU Depth V2 dataset~\cite{silberman2012indoor} and FlyingThings3D dataset~\cite{mayer2016large}.
While NYUv2 serves as the standard real-world indoor benchmark, FlyingThings3D contributes large-scale synthetic data with high geometric and textural diversity, enriching the training distribution.
To model the physical sensor geometry, we discretize the curved focal surface using an approximately $6000\times4500$ numerical grid with an equivalent projected sampling pitch of $2~\mu\mathrm{m}$. These values define the nominal sampling specification used for sensor--optics co-design and forward simulation.

Based on this physical model, we organize the datasets into distinct training and evaluation protocols.
For training, we use the official NYUv2 training split together with FlyingThings3D. Each NYUv2 image is mapped to an approximately $12^\circ$ local FOV.
Each sample is treated as a local tangent plane to the spherical image surface. At the corresponding half angle of $6^\circ$, the relative difference between $\tan\alpha$ and $\alpha$ is approximately $0.37\%$, indicating a small local projection discrepancy.
The NYUv2 samples and synthetic patches from FlyingThings3D are randomly placed at different sensor positions to sample the corresponding spatially varying PSFs.
This protocol evaluates reconstruction under optical degradation that varies with field position across the $120^\circ$ optical FOV; it does not represent a single geometrically coherent $120^\circ$ scene.
For evaluation, quantitative results are reported on the NYUv2 test split, while the Middlebury dataset~\cite{scharstein2014high} is used to assess restoration near the field edge.
The high-resolution Middlebury images ($\sim3000\times2000$) are mapped over a wider numerical sensor region spanning approximately $60^\circ$ and placed near the field edge to assess restoration under strongly varying peripheral PSFs. This setting does not represent a geometrically coherent $60^\circ$ spherical scene.

\noindent\textbf{Implementation Details.} 
All experiments are conducted on a single NVIDIA $A800$ GPU. 
The reconstruction network is trained using the Adam optimizer~\cite{kingma2014adam} with hyperparameters set to ${\beta _1}{=}0.9$ and ${\beta _2} {=} 0.99$. 
We employ an initial learning rate of $5e{-}4$ and a batch size of $8$. 
To improve model generalization, we apply data augmentation techniques, including random horizontal flips and rotations. During training, randomly cropped $256{\times}256$ patches are utilized as inputs. 
To balance the contributions of the image restoration and depth estimation tasks, the weights for the content loss ${L_{cont}}$, the frequency-domain loss ${L_{MSFR}}$, and the Silog loss ${L_{Silog}}$ are set to $1.0, 0.1, 0.1$, respectively. 
The model was trained for a total of $300,000$ iterations.
For a $480\times640$ input, the network contains $10.08$ M parameters and requires $942.58$ GFLOPs. On an NVIDIA A800 GPU, the average FP32 inference time is $49.46~\mathrm{ms}$ ($20.22$ FPS), with a peak allocated memory of $786.84~\mathrm{MB}$. Full-resolution deployment would require tiled inference and further implementation optimization.

\subsection{Comparative Evaluation and Benchmark Analysis}

\begin{table}[!htb]\scriptsize
    \centering
       \caption{Quantitative comparison of joint image restoration and depth estimation on the NYU Depth V2 dataset.}
       \label{tab:Quantitative_comparsion}  
\renewcommand{\arraystretch}{1.2} 
\resizebox{1.0\linewidth}{!}{
\begin{tabular}{lccccccccc}
    \toprule 
    \multicolumn{1}{c}{Method} & \multicolumn{3}{c}{Depth Accuracy$\uparrow$, $\delta <$} & \multicolumn{2}{c}{Depth Error$\downarrow$} & \multicolumn{2}{c}{Image Error} & \multicolumn{2}{c}{Others} \\
    \midrule 
    & $1.25$ & $1.25^2$ & $1.25^3$ & RMSE & Abs Rel  & PSNR(dB)$\uparrow$ & SSIM$\uparrow$ &Range(m) & Noise \\
    \cmidrule(lr){2-4} \cmidrule(lr){5-6} \cmidrule(lr){7-8} \cmidrule(l){9-10}
    ZoeDepth~\cite{bhat2023zoedepth}  & 0.955 & 0.995 & 0.999 & 0.270 & 0.075  &- &-  &0.8-10 &- \\
    VPD~\cite{zhao2023unleashing}     & 0.964 &0.995  &0.999  & 0.254 & 0.069  &- &-  &0.8-10 &- \\
    ECoDepth~\cite{patni2024ecodepth} & 0.978 & 0.997 &0.999  & 0.218 & 0.048  &- &-  &0.8-10 &- \\
    Depthanything~\cite{yang2024depth}& 0.984 & 0.998 & 1.000 & 0.206 & 0.056  & - & - &0.8-10 & - \\
    Metric3Dv2~\cite{hu2024metric3d}  & 0.989 & 0.998 & 1.000 & 0.180 & 0.046  & - & - &0.8-10 & - \\
    \midrule 
    DeepOptics~\cite{chang2019deep}        &0.930 &0.990 &0.999 &0.433 &0.087 &- &- &0.5-8.0 &-\\
    Phase3D~\cite{wu2019phasecam3d}        &0.832 &0.989 &0.997 &0.382 &0.093 &- &- &0.4-1.0 &0.01\\
    Learnedoptics~\cite{ikoma2021depth}   &0.959 &0.990 &0.996 &0.439 &0.070 &28.54 &- &1.0-5.0 &0.005\\
    CF-DOE~\cite{zhuge2024calibration}  & 0.987 & 0.998 & 1.000 & 0.224 & 0.044 & 29.36 & 0.912 & 0.8-10 & 0.005 \\
    Off-aperture DOE~\cite{wei2025learned}$^{\dagger}$& 0.813 & 0.974 & 0.996& 0.656 & 0.139& 22.77 & 0.677& 0.8--10 & 0.003 \\
    Split-Aperture~\cite{shi2024split}  & 0.988 & 0.998 & 1.000 & 0.169 & 0.046  & - & - & 1.0-5.0 & 0.01 \\
    \midrule 
   Ours (Ideal) & 0.992 & 0.999  &1.000  & 0.161  & 0.035  &31.15 &0.945 & 0.8-10 &- \\
    Ours (Standard)     & 0.988  &0.999  &1.000  &0.205  &0.044   &30.57  & 0.938  &0.8-10  &0.005 \\

    \bottomrule 
\end{tabular}%
}
\parbox{\linewidth}{
\footnotesize
$^\dagger$System-level reproduction based on the publicly described thin-lens DOE configuration; the compound-lens prescription and precomputed optical assets were unavailable.
}

\end{table}

\begin{figure}[!htb]
    \centering
    \includegraphics[width=1.0\linewidth]{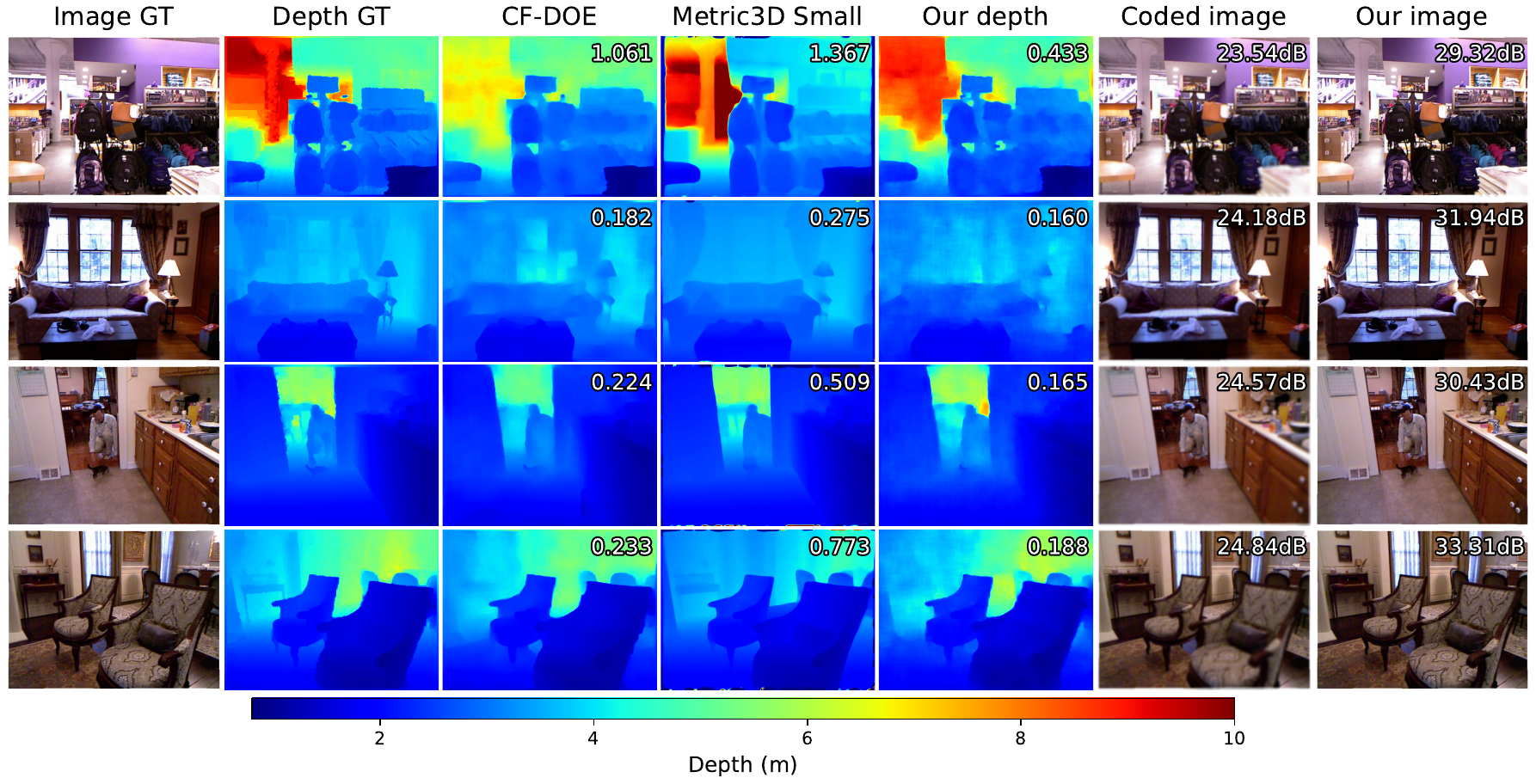}
    \caption{Qualitative comparison on NYU Depth V2. 
    We benchmark against CF-DOE~\cite{zhuge2024calibration} and Metric3D Small~\cite{hu2024metric3d}. Inset values denote Depth RMSE (m) and Image PSNR (dB). Observe that our standard method yields reduced artifacts and superior fidelity.}
    \label{fig:Qualitative_nyu}
\end{figure}

\begin{figure}[!htb]
    \centering
    \includegraphics[width=1.0\linewidth]{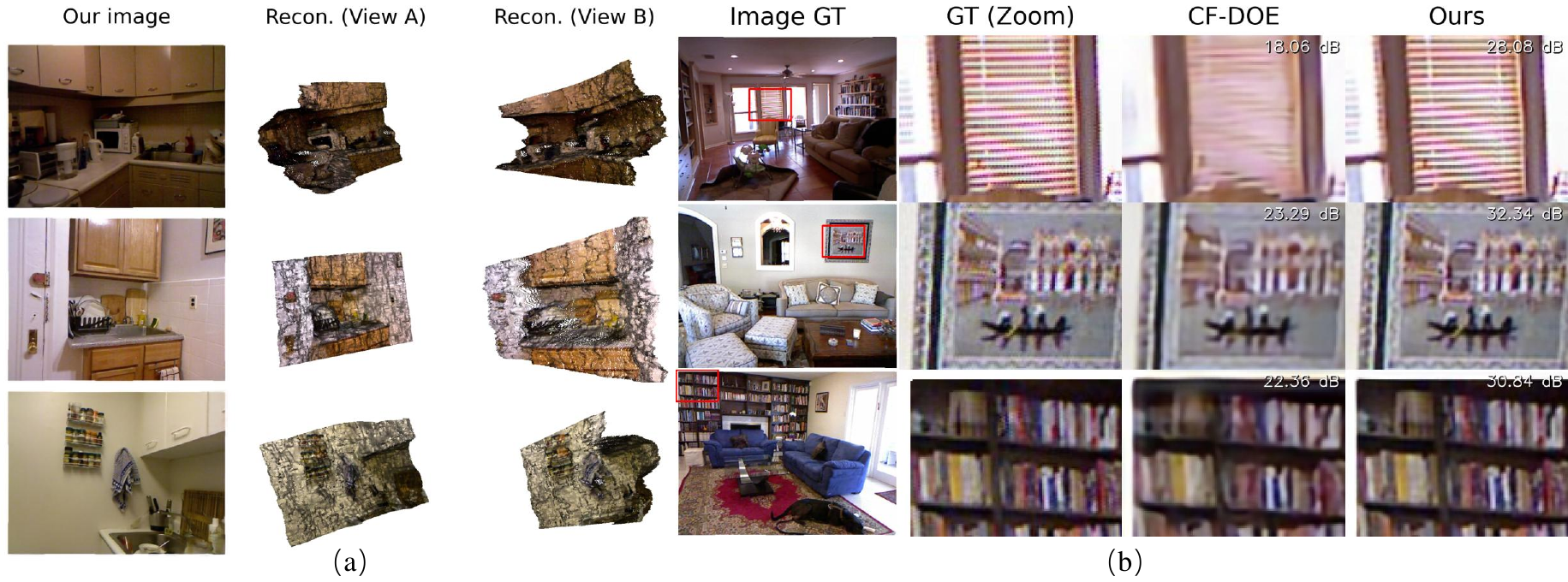}
    \caption{Qualitative evaluation on the NYU Depth V2 dataset. (a) Qualitative 3D reconstruction results. (b) Enlarged qualitative comparison for image restoration.}
    \label{fig:recon_enlarge}
\end{figure}

\noindent\textbf{Comparative performance analysis.}
To quantitatively assess performance, we employ standard metrics: Absolute Relative Error (Abs Rel), Root Mean Square Error (RMSE), and accuracy thresholds ($\delta < 1.25^n$) for depth evaluation; and Peak Signal-to-Noise Ratio (PSNR) alongside Structural Similarity Index Measure (SSIM)~\cite{wang2004image} for image restoration. Consistent with standard benchmarks, all metrics are computed strictly on valid pixels.

As detailed in Table~\ref{tab:Quantitative_comparsion}, we conduct a stratified comparative analysis to rigorously benchmark the proposed BMI framework against established paradigms.
To ensure a strictly fair comparison against baselines inherent to standard FOV, we restrict our quantitative evaluation to the central FOV of the NYUv2 dataset for all competing methods. 
First, we benchmark against software-only Monocular Depth Estimation (MDE) baselines. It is noteworthy that while MDE methods benefit from ideal, all-in-focus inputs, our method achieves superior metric accuracy even when decoding from optically blurred measurements.
Quantitatively, our ideal system achieves an Abs Rel of $0.035$ and an RMSE of $0.161$, distinctly outperforming leading foundation models such as Metric3Dv2~\cite{hu2024metric3d} ($0.046$ Abs Rel, $0.180$ RMSE).
This demonstrates that the deterministic aberration cues encoded by our monocentric optics provide a tangible physical advantage over purely algorithmic inference.

The second tier of evaluation benchmarks our integrated system against comprehensive deep optics frameworks. 
Compared to the diffractive encoding strategy of CF-DOE~\cite{zhuge2024calibration}, our refractive design achieves higher image-restoration performance under the reported evaluation settings. While diffractive elements often introduce spectral dispersion that degrades image restoration (PSNR $29.36$ dB), our standard system, all-spherical approach preserves spectral consistency, achieving a significantly higher PSNR of $30.57$ dB. Moreover, our continuous aberration encoding proves more robust than discrete diffractive modulation for depth precision, reducing the RMSE from $0.224$ to $0.205$.
Finally, we benchmark against the state-of-the-art Split-Aperture scheme~\cite{shi2024split}. Quantitatively, under standard conditions, our method achieves a superior Abs Rel ($0.044$ \textit{vs.} $0.046$) and a comparable RMSE ($0.205$). Furthermore, in the ideal setting, our system demonstrates clear superiority with an RMSE of $0.161$. 
Critically, while Split-Aperture relies on dual-channel hardware redundancy (simultaneous focused and defocused captures) effectively constraining it to a limited depth interval ($1.0{\sim}5.0$ m) and standard FoV, our system solves the more challenging single-shot blind deconvolution problem.  This enables robust encoding across a significantly extended range ($0.8{\sim}10.0$ m) and a hemispherical $120^{\circ}$ field, confirming the superiority of our bio-inspired design without requiring complex external masks.
A further input-scale ablation shows that removing the two coarse-scale branches decreases the restoration PSNR from $31.15$ to $30.60~\mathrm{dB}$ and increases the depth RMSE from $0.161$ to $0.177~\mathrm{m}$. 
The planar sensor variant shows substantial degradation even over a restricted field range, while suppressing spherical aberration increases the depth RMSE from $0.191$ to $0.329~\mathrm{m}$. Complete settings, results, and comparison limitations are provided in Table~S5.

\noindent\textbf{Qualitative result analysis.}
Figure~\ref{fig:Qualitative_nyu} presents the qualitative results on the NYUv2 dataset. 
Our method recovers depth maps with superior structural fidelity, effectively delimiting object boundaries. 
While the inherent ill-posedness of deconvolution may introduce minor ringing artifacts at sharp discontinuities, our approach significantly outperforms baselines like Metric3D~\cite{hu2024metric3d} in overall geometric consistency.
Crucially, although depth estimation accuracy naturally follows an optical attenuation law for distant objects (due to reduced PSF sensitivity in the far-field), our physics-driven decoding still yields the most reliable results in these challenging regions. This effectively mitigates the severe absolute scale ambiguity that typically plagues pure software-based MDE approaches.

We further extend this analysis to 3D scene recovery and texture restoration in Figure~\ref{fig:recon_enlarge}. The left panel (a) visualizes the 3D point cloud reconstruction from two distinct viewpoints, demonstrating that our system successfully acquires both high-fidelity RGB images and valid metric depth simultaneously. This result validates the joint RGBD reconstruction capability of the proposed framework under simulated field-dependent optical degradation.
The right panel (b) provides a detailed qualitative comparison of image restoration performance. As evidenced by the magnified regions, our method exhibits exceptional capability in recovering high-frequency spatial details. This is particularly distinct in areas with complex textures, such as window blinds, the hanging painting, and the bookcase. Compared to competing methods, our approach more effectively restores the clarity and sharpness of the image, significantly reducing restoration artifacts.

\subsection{Large Field Fidelity and Signal Robustness Analysis}

\begin{table}[!htb]\scriptsize
    \centering
       \caption{Quantitative evaluation of wide-field consistency on Middlebury. We compare depth metrics against MDE methods and a regular fisheye baseline.}
       \label{tab:middlebury_comparsion}  
\renewcommand{\arraystretch}{1.2} 
\resizebox{0.8\textwidth}{!}{%
\begin{tabular}{lcccccc}\hline
    \toprule %
    \multicolumn{1}{c}{Method} & \multicolumn{3}{c}{Depth Accuracy$\uparrow$, $\delta <$} & \multicolumn{2}{c}{Depth Error$\downarrow$}\\
    \midrule 
    & $1.25$ & $1.25^2$ & $1.25^3$ & RMSE & Abs Rel \\
    \cmidrule(lr){2-4} \cmidrule(lr){5-6}
    Metric3Dv2-small~\cite{hu2024metric3d}   &0.217 &0.658 &0.809 &1.700 &0.588\\
    Metric3Dv2-giant~\cite{hu2024metric3d}   &0.221 &0.851 &0.935 &1.306 &0.424\\
    DepthAnythingv2-small~\cite{yang2024depth}   &0.219 &0.510 &0.782 &1.143 &0.358\\
    DepthAnythingv2-large~\cite{yang2024depth}   &0.146 &0.445 &0.785 &1.182 &0.382\\
    Regular fisheye  &0.295 &0.553 &0.746 &1.806 &0.539 \\
    Ours (Standard)  &0.805 &0.958 &0.993 &0.590 &0.128\\
    \bottomrule
\end{tabular}%
}
\end{table}

\begin{figure}[!htb]
    \centering
    \includegraphics[width=0.95\linewidth]{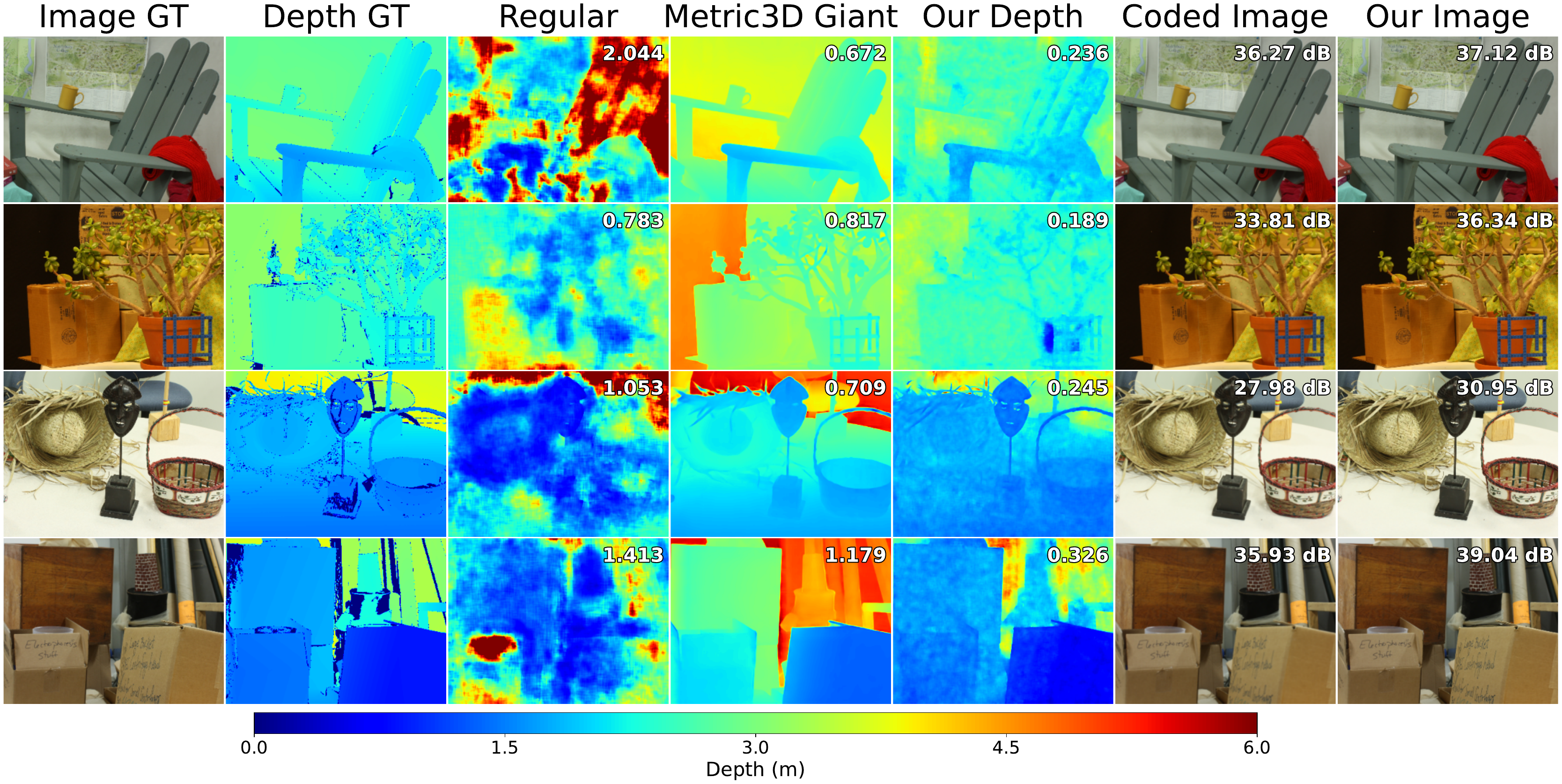}
    \caption{Qualitative comparison on Middlebury. Inset values denote the Depth RMSE (m) and Image PSNR (dB).}
    \label{fig:Qualitative_middlebury}
\end{figure}

\noindent\textbf{Field-position consistency analysis.}
While standard benchmarks provide a baseline, they often lack sufficient spatial resolution required to rigorously test optical fidelity at the extreme edges of a large field.
To validate the off-axis consistency of our system, we employ the high-resolution Middlebury dataset~\cite{scharstein2014high} and map the samples to representative peripheral sensor positions to evaluate joint reconstruction under challenging field-dependent PSFs.
As quantified in Table~\ref{tab:middlebury_comparsion}, our method demonstrates strong zero-shot performance at these sampled field positions (where models are not trained on Middlebury), achieving an RMSE of $0.590$ and Abs Rel of $0.128$, distinctly outperforming state-of-the-art MDE algorithms.

To provide a system-level comparison between the optical front ends, we construct a conventional fisheye baseline using the same physically based simulation framework and reconstruction network architecture. The network is independently retrained using the spatially varying PSFs generated by the fisheye design. As reported in Table~\ref{tab:middlebury_comparsion}, the fisheye baseline yields a depth RMSE of $1.806~\mathrm{m}$, whereas BMI achieves $0.590~\mathrm{m}$ under the same evaluation protocol. This trend is consistent with the CRLB analysis in Sec.~\ref{section:PSF analysis}, which indicates weaker theoretical depth sensitivity for the selected conventional fisheye design. Because the two systems differ in both aberration structure and sensor geometry, this comparison is interpreted as a system-level baseline rather than a component attribution.

Visually, Figure~\ref{fig:Qualitative_middlebury} corroborates these findings. While the foundation model Metric3Dv2 yields visually coherent and smooth depth maps by leveraging its vast semantic priors and explicitly provided camera intrinsics, it suffers from severe absolute scale drift.
For instance, as shown in the first row, Metric3Dv2 erroneously extrapolates the background, hallucinating a spatial depth that contradicts physical reality. This highlights the inherent limitation of data-driven priors with unfamiliar focal geometries.
Meanwhile, the fisheye baseline fails to recover structure entirely. In contrast, our method resolves the correct absolute distance. This validates that our structured aberrations function as a deterministic physical anchor, enabling a standard network to achieve substantially more accurate metric-depth recovery than the evaluated conventional fisheye baseline under the same reconstruction protocol.

\begin{table}[!htb]\scriptsize
    \centering
       \caption{Impact of sensor noise levels on wide-field image restoration and depth estimation accuracy evaluated on the Middlebury benchmark.}
       \label{tab:compared_with_noise}  
\renewcommand{\arraystretch}{1.2} 
\resizebox{0.8\textwidth}{!}{%
\begin{tabular}{lcccccccc}\hline
    \toprule %
    \multicolumn{1}{c}{Noise} & \multicolumn{3}{c}{Depth Accuracy$\uparrow$, $\delta <$} & \multicolumn{2}{c}{Depth Error$\downarrow$} & \multicolumn{2}{c}{Image Error$\uparrow$}\\
    \midrule 
    & $1.25$ & $1.25^2$ & $1.25^3$ & RMSE & Abs Rel & PSNR(dB) & SSIM \\
    \cmidrule(lr){2-4} \cmidrule(lr){5-6} \cmidrule(lr){7-8}
    0       &0.862 &0.958 &0.983 &0.650 &0.115 &34.88 &0.951\\
    0.005   &0.805 &0.958 &0.993 &0.590 &0.128 &34.59 &0.946\\
    0.010   &0.720 &0.912 &0.972 &0.859 &0.185 &34.39 &0.942\\
    0.020   &0.674 &0.870 &0.948 &0.886 &0.198 &34.19 &0.937\\
    0.050   &0.489 &0.730 &0.874 &1.256 &0.295 &33.70 &0.924\\
    \bottomrule
\end{tabular}%
}
\end{table}

\noindent\textbf{Noise sensitivity analysis.}
Depth-from-defocus algorithms are intrinsically sensitive to the signal-to-noise ratio (SNR). To determine operational limits, we conduct a stress test by injecting additive Gaussian noise with standard deviations up to $\sigma=0.05$ into the Middlebury simulation. Such a high variance corresponds to nearly $13$ gray levels in an 8-bit domain and poses a rigorous challenge to signal recovery.
As detailed in Table~\ref{tab:compared_with_noise}, while depth RMSE naturally correlates with the noise floor, the system exhibits graceful degradation rather than catastrophic failure. Even when the noise level increases tenfold from the baseline $\sigma=0.005$ to a challenging $\sigma=0.05$, the RMSE remains constrained (increasing from $0.590$ to $1.256$).
This indicates that the depth-encoding aberration rings function as robust, spatially distributed signal carriers. Since the depth information is embedded within the substantial spatial footprint of the PSF rather than localized intensity spikes, the network can effectively integrate signals across the pattern to suppress stochastic pixel-level noise.

\noindent\textbf{Robustness to representative hardware errors.}
To evaluate mismatch between the hardware and the calibrated model, we consider a $0.5$ pixel PSF centroid shift, random PSF scale errors of $3\%$ and $5\%$, a $10~\mu\mathrm{m}$ lens decenter, a $1$ arcmin lens tilt, $3\%$ diffuse stray light contamination, and their combined effect. The decenter and tilt are applied to the complete monocentric lens assembly. All perturbed measurements are evaluated using the fixed pretrained network without retraining.

\begin{table}[!htb]\scriptsize
    \centering
    \caption{Robustness under representative hardware errors on NYU Depth V2.}
    \label{tab:hardware_error}
    \renewcommand{\arraystretch}{1.15}
    \resizebox{0.8\linewidth}{!}{
    \begin{tabular}{lcccc}
        \toprule
        Test condition
        & RMSE (m)$\downarrow$ & Abs Rel$\downarrow$ & PSNR (dB)$\uparrow$ & SSIM$\uparrow$ \\
        \midrule
        Nominal
        & 0.161 & 0.035 & 31.151 & 0.945 \\
        PSF shift, $0.5$ pixel
        & 0.191 & 0.039 & 27.562 & 0.917 \\
        PSF scale error, $3\%$
        & 0.175 & 0.037 & 31.120 & 0.945 \\
        PSF scale error, $5\%$
        & 0.196 & 0.040 & 31.080 & 0.944 \\
        Lens tilt, $1$ arcmin
        & 0.162 & 0.035 & 31.230 & 0.946 \\
        Lens decenter, $10~\mu\mathrm{m}$
        & 0.322 & 0.066 & 30.959 & 0.942 \\
        Diffuse stray light, $3\%$
        & 0.182 & 0.040 & 30.700 & 0.939 \\
        Joint perturbation
        & 0.287 & 0.061 & 27.406 & 0.907 \\
        \bottomrule
    \end{tabular}%
    }
    \parbox{\linewidth}{
    \footnotesize
    All perturbed cases were evaluated using the fixed pretrained network without retraining. The joint condition includes a $0.5$-pixel PSF shift, a $3\%$ random PSF error, a $10~\mu\mathrm{m}$ decenter, a $1$-arcmin tilt, and $3\%$ veiling glare.
    }
\end{table}

As shown in Table~\ref{tab:hardware_error}, the network is relatively insensitive to a $1$ arcmin tilt and moderate PSF scale errors. A $0.5$ pixel PSF shift has a larger influence on image restoration, reducing the PSNR from $31.151$ to $27.562~\mathrm{dB}$. The $10~\mu\mathrm{m}$ decenter produces the largest individual depth error, increasing the RMSE from $0.161$ to $0.322~\mathrm{m}$. Under the joint perturbation, the RMSE is $0.287~\mathrm{m}$ and the Abs Rel is $0.061$. 
These simulations suggest that lens decenter may require particular attention during assembly and provide initial sensitivity estimates for future prototype design and calibration.

\noindent\textbf{Manufacturability context and sensor radius sensitivity.}
The curved sensor adopted in this work defines a target design point for the proposed sensor-optics architecture. Existing experimental studies provide complementary evidence for its practical realization. Commercial $18$-megapixel BSI CMOS sensors with a $1.25~\mu\mathrm{m}$ pixel pitch have been thinned, spherically formed to radii of $16.7$--$24.6~\mathrm{mm}$, packaged, and integrated into functional cameras \cite{guenter2017highly}. 
At higher curvatures, silicon photodetector arrays have been demonstrated on surfaces with radii of approximately $7.2$--$7.5~\mathrm{mm}$ \cite{zhang2017origami,choi2024anti}.
In addition, an ultrathin complete CMOS imager has been wire-bonded, packaged, and operated for video imaging on a cylindrical surface with a radius of $20~\mathrm{mm}$ \cite{imura2024demonstration}. Curved CMOS sensor packaging and preliminary environmental reliability testing have also been reported~\cite{joaquina2023curved}.

To the best of our knowledge, the studies reviewed above do not report a single sensor that combines a spherical radius near $7.2~\mathrm{mm}$, approximately $6000\times4500$ addressable pixels, the sampling density corresponding to the equivalent projected pitch of $2~\mu\mathrm{m}$ used in our model, and single exposure RGB imaging.
We therefore treat this sensor configuration as a forward-looking target for joint sensor and optics development rather than as a commercially available component.

Table~\ref{tab:fabrication tolerance} evaluates sensitivity to a uniform perturbation of the global spherical radius. At a $0.5\%$ perturbation, the depth RMSE increases from $0.590$ to $0.813~\mathrm{m}$, while the PSNR changes by less than $0.1~\mathrm{dB}$. 
At a $2\%$ perturbation, the RMSE reaches $1.659~\mathrm{m}$ and the PSNR decreases by $0.66~\mathrm{dB}$. These results show that image restoration is relatively insensitive to modest global radius mismatch, whereas depth estimation is more sensitive. The analysis provides an initial quantitative sensitivity reference for fabrication and calibration, but does not model local surface figure errors or qualify the manufacturability of the complete sensor specification.

\begin{table}[!htb]\scriptsize
    \centering
       \caption{Impact of sensor curvature deviation on joint image restoration estimation and depth estimation performance evaluated on the Middlebury benchmark.}
       \label{tab:fabrication tolerance}  
\renewcommand{\arraystretch}{1.2} 
\resizebox{0.8\textwidth}{!}{%
\begin{tabular}{lccccccc}\hline
    \toprule %
    \multicolumn{1}{c}{Sensor-radius perturbation} & \multicolumn{3}{c}{Depth Accuracy$\uparrow$, $\delta <$} & \multicolumn{2}{c}{Depth Error$\downarrow$} & \multicolumn{2}{c}{Image Error$\uparrow$}\\
    \midrule 
    & $1.25$ & $1.25^2$ & $1.25^3$ & RMSE & Abs Rel & PSNR(dB) & SSIM \\
    \cmidrule(lr){2-4} \cmidrule(lr){5-6} \cmidrule(lr){7-8}  
    0       &0.805 &0.958 &0.993 &0.590 &0.128 &34.59 &0.946\\
    0.5\% ($\approx36\mu m$)   &0.774 &0.940 &0.974 &0.813 &0.170 &34.67 &0.944\\
    1\% ($\approx72\mu m$)     &0.603 &0.846 &0.939 &1.112 &0.260 &34.57 &0.939\\
    2\% ($\approx144\mu m$)      &0.375 &0.639 &0.801 &1.659 &0.465 &33.93 &0.929\\
    \bottomrule
\end{tabular}%
}
\end{table}

\section{Conclusion and Future Work}
In conclusion, to resolve the fundamental physical conflict between compact form factors and high-fidelity 3D sensing in large-field-of-view applications, our paper proposes the Bio-inspired Monocentric Imaging (BMI) framework.
Unlike traditional designs that suppress aberrations, our holistic co-design strategy actively exploits intrinsic aberration as a deterministic depth carrier. 
By adopting a monocentric topology, the system suppresses asymmetric off-axis aberrations and retains structured depth encoding at field positions spanning the optical FOV.

Validated by both Cramér-Rao Lower Bound analysis and full-field wave optics simulations, our method demonstrates superior performance limits compared to conventional designs. 
On NYU Depth V2, the framework achieves a depth RMSE of $0.161 m$ and an image PSNR of $31.15 dB$, significantly outperforming state-of-the-art software-only foundation models. Field-position evaluations further demonstrate consistent reconstruction under spatially varying PSFs sampled across the $120^{\circ}$ optical FOV.
Beyond these metrics, our results support a central principle in computational imaging: meticulously designing the optical front-end to optimize information quality at the physical source offers a promising complement to increasingly complex algorithmic inference.

In this study, the system is evaluated primarily through a physically based forward model because realization of the specified curved RGB sensor requires dedicated joint development of the sensor and optics. Existing experiments separately demonstrate several relevant enabling capabilities, including high-density spherical CMOS forming, ultrathin CMOS operation, highly curved silicon photodetector arrays, and curved sensor packaging. The sensor radius sensitivity, broadband spectral, and representative hardware perturbation analyses provide initial quantitative references for future fabrication, alignment, calibration, and prototype integration. Future work will focus on prototype integration and measured PSF calibration to evaluate the BMI framework experimentally.

\section*{Funding}
The Henan Province Key R\&D Special Project (231111112700);
The National Natural Science Foundation of China (Grant No. 62473139);
The Hunan Provincial Research and Development Project (Grant No. 2025QK3019); The State Key Laboratory of Autonomous Intelligent Unmanned Systems (the opening project number ZZKF2025-2-10).

\section*{Disclosures}
The authors declare no conflicts of interest.

\section*{Data availability}
The source code and reproducibility materials underlying the results presented in this paper are publicly available at \url{https://github.com/ZongxiYu-ZJU/BMI}. The public datasets used in this study are available from their original sources cited in the manuscript.

\bibliography{sample}

\end{document}